\documentclass[runningheads, envcountsame, a4paper]{llncs}
\usepackage[pdftex]{graphicx}
\usepackage{epstopdf}
\usepackage[hyphens]{url}
\usepackage{microtype}
\usepackage[misc]{ifsym}
\usepackage{graphicx}
\usepackage{times}
\usepackage{soul}
\usepackage[utf8]{inputenc}
\usepackage{amsmath}
\usepackage{amsfonts}
\usepackage{booktabs}
\usepackage{algorithm}
\usepackage{algorithmic}
\usepackage{bm}
\usepackage{comment}
\usepackage{array}
\usepackage[symbol]{footmisc}
\begin{document}
\title{AutoGraph: Automated Graph Neural Network}
%
%
\author{Yaoman Li\inst{1,2}\Letter 
\and Irwin King\inst{1}}
%
%
\institute{Department of Computer Science \& Engineering \\ The Chinese University of Hong Kong \\ Shatin, NT, Hong Kong
\and Lenovo Machine Intelligence Center, Cyberport, Hong Kong 
\email{$\{$ymli,king$\}$@cse.cuhk.edu.hk}}

\toctitle{AutoGraph: Automated Graph Neural Network}
\tocauthor{Yaoman~Li and Irwin~King}
\authorrunning{Yaoman Li and Irwin King}
\maketitle              
\setcounter{footnote}{0}

\begin{abstract}
Graphs play an important role in many applications. Recently, Graph Neural Networks (GNNs) have achieved promising results in graph analysis tasks. Some state-of-the-art GNN models have been proposed, e.g., Graph Convolutional Networks (GCNs), Graph Attention Networks (GATs), etc. Despite these successes, most of the GNNs only have shallow structure. This causes the low expressive power of the GNNs. To fully utilize the power of the deep neural network, some deep GNNs have been proposed recently. However, the design of deep GNNs requires significant architecture engineering. In this work, we propose a method to automate the deep GNNs design. In our proposed method, we add a new type of skip connection to the GNNs search space to encourage feature reuse and alleviate the vanishing gradient problem. We also allow our evolutionary algorithm to increase the layers of GNNs during the evolution to generate deeper networks. We evaluate our method in the graph node classification task. The experiments show that the GNNs generated by our method can obtain state-of-the-art results in Cora, Citeseer, Pubmed and PPI datasets.

\end{abstract}
\section{Introduction}

Graph Neural Networks (GNNs) are deep learning-based methods that have been successfully applied in graph analysis. It is one of the most important machine learning tools for solving graph problems. Unlike other machine learning data, graphs are non-Euclidean data. Many real-world problems can be modeled as graphs, such as knowledge graphs, protein-protein interaction networks, social networks, etc. The neural networks like Recurrent Neural Networks (RNNs) or Convolutional Neural Networks (CNNs) cannot directly apply to graph data. Hence, GNNs have received more and more attention. Some GNN models have been proposed and obtain promising results on some graph tasks, such as node classification \cite{DBLP:conf/iclr/KipfW17,DBLP:conf/nips/HamiltonYL17,DBLP:conf/iclr/VelickovicCCRLB18,DBLP:journals/pr/ManessiRM20}, link prediction \cite{DBLP:conf/ijcai/ZhangSZK19} and clustering \cite{DBLP:conf/nips/YingY0RHL18}. 

However, most of the GNNs suffer the low expressive power problem due to their shallow architectures. Some works \cite{DBLP:conf/nips/LuanZCP19,DBLP:journals/corr/abs-1908-05081} have been proposed to solve this problem. The design of deep GNNs requires a huge amount of human effort for neural architecture tuning. GNN models are usually very sensitive to the hyperparameters, for different tasks, we might also need to adjust the hyperparameters to obtain the optimal result. For example, the activation function needs to be carefully selected to avoid features degradation \cite{DBLP:conf/nips/LuanZCP19}, the number of attention heads of GAT \cite{DBLP:conf/iclr/VelickovicCCRLB18} needs to be carefully selected for different data, etc. The variants of GNNs may have a better performance in some specific problems. It is impossible to explore all possibilities manually. 

We notice that the Neural Architecture Search (NAS) has archived great success in designing the CNNs and RNNs for many computer vision and language modeling tasks \cite{DBLP:conf/iclr/ZophL17,DBLP:conf/icml/RealMSSSTLK17,DBLP:conf/isnn/LiK19}. Many NAS methods for CNNs and RNNs have been proposed recently. For example, Zoph et al. \cite{DBLP:conf/iclr/ZophL17} apply reinforcement learning to design CNNs for image classification problems. They use a recurrent network controller to generate CNN models and use the validation result of the CNN models as a reward to update the controller. Real et al. \cite{DBLP:conf/icml/RealMSSSTLK17} design an evolutionary algorithm to evolve the CNN models from scratch and obtain state-of-the-art results. However, these works cannot be applied to GNNs directly.

Inspired by the success of NAS in designing CNNs and RNNs, recent works \cite{DBLP:journals/corr/abs-1904-09981,DBLP:journals/corr/abs-1909-03184} are tried to apply NAS methods to design GNN models for citation networks. They propose to use reinforcement learning to design the GNN models. However, their proposed method can only generate fixed-length GNN models, and the generated GNN models only have shallow architectures. The deep GNNs generated by their methods will suffer the over-smoothing problem.

\begin{figure}[tb]
    \centering
    \includegraphics[width=0.9\textwidth]{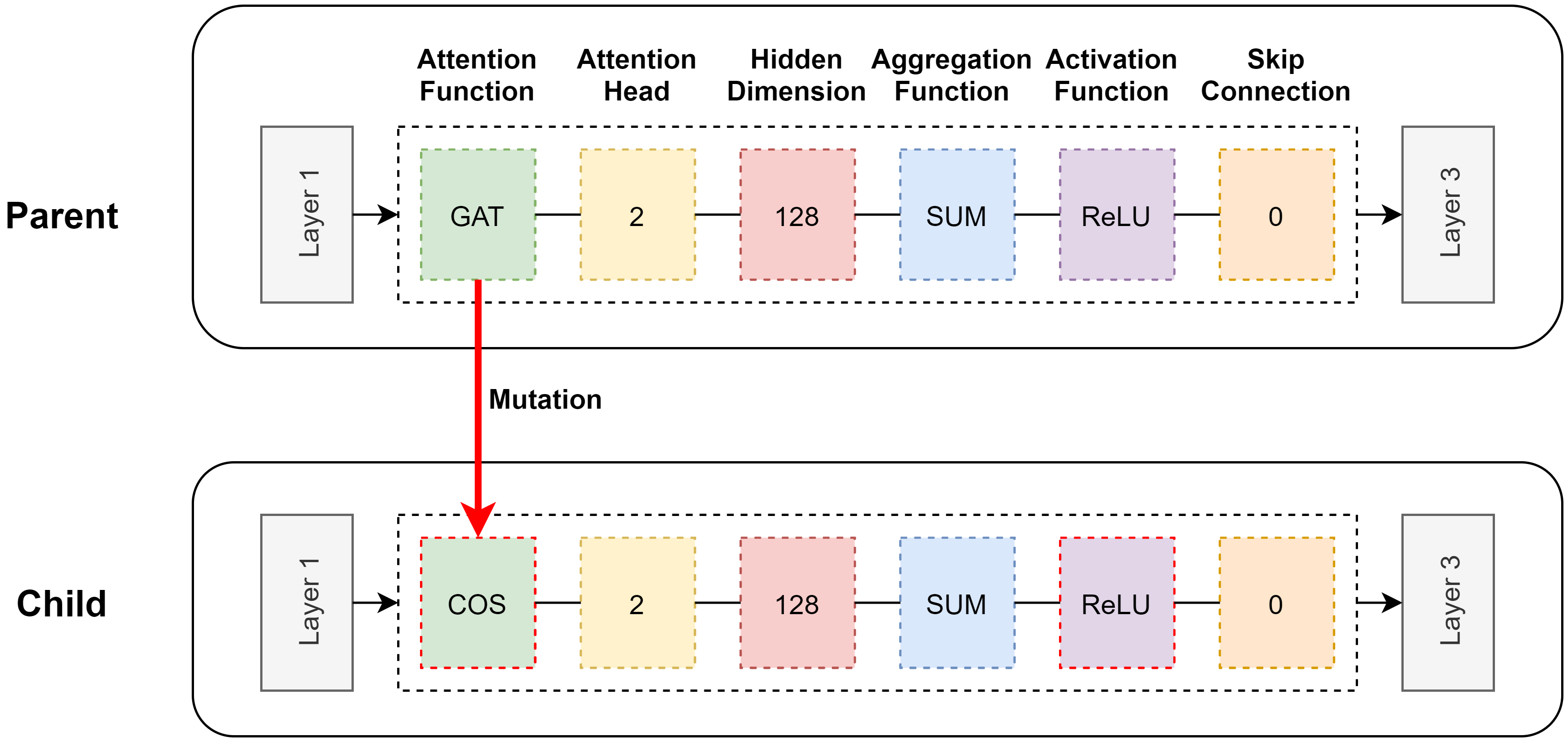}
    \caption{Graph neural network architecture evolution example. The GNN architecture can be encoded by six states, i.e., Attention Function, Attention Head, Hidden Dimension, Aggregation Function, Activation Function and Skip Connection.}
    \label{fig:mutation}
\end{figure}

To overcome the above-mentioned problem, we propose a new AutoGraph method that applies an evolutionary algorithm to automatically generate deep GNNs. We first design a new search space and schema for the GNN model, which allows GNN with various layers and covers most of the state-of-the-art models. Then we apply evolutionary algorithm and mutation operations to evolve the initial GNN models. Next, we demonstrate a method to search for the best hyperparameters for the new GNN models which allow us to fairly compare the generated models and improve the robustness of our method. Finally, we conduct experiments on both transductive and inductive learning tasks and compare our method with baseline GNNs and the models generated by other reinforcement learning and random search strategies. The results show that we can generate state-of-the-art models for all test data efficiently. In summary, our contributions are:

\begin{itemize}
    \item To the best of our knowledge, we are the first to study deep GNNs by using NAS. Our method can automate the architecture engineering process for deep GNNs, which can save many human efforts.
    \item Experiment results show that our proposed method can search for deep GNN models for different tasks efficiently.
    \item The GNN models generated by our method can outperform the handcrafted state-of-the-art GNN models.
\end{itemize}

\section{Related Work}

Inspired by CNNs \cite{lecun1998gradient,DBLP:conf/nips/KrizhevskySH12} and graph embedding \cite{DBLP:journals/tkde/CuiWPZ19,DBLP:conf/www/0004ZMK20}, GNNs are proposed to collectively aggregate information from graph structure. It is first proposed in \cite{DBLP:journals/tnn/ScarselliGTHM09}. GNNs have been widely applied for graph analysis \cite{DBLP:journals/corr/abs-1812-08434,DBLP:journals/corr/abs-1901-00596} recently. The target of GNNs is to learn a representation of each node $\mathbf{h}_v \in \mathbb{R}^s$ which contains information for its neighborhood. The $\mathbf{h}_v$ also called a state embedding of a node. It can be used to produce an output $\mathbf{o}_v$, e.g., the node labels. They can defined as follows \cite{DBLP:journals/corr/abs-1812-08434}:
\begin{align}
    \mathbf{h}_v &= f(\mathbf{x}_v,\mathbf{x}_{co[v]},\mathbf{h}_{ne[v]},\mathbf{x}_{ne[v]}),\\
    \mathbf{o}_v &= g(\mathbf{h}_v,\mathbf{x}_v),
\end{align}
where $f$ is the transition function that updates the node state according to the neighborhood, $g$ is the output function that generates output from the node state and features. $\mathbf{x}_v$,$\mathbf{x}_{co[v]}$,$\mathbf{x}_{ne[v]}$,$\mathbf{h}_{ne[v]}$ are the features of $v$, the features of its edges, the features and the states of its neighborhood, respectively.

Let $\mathbf{H}$, $\mathbf{O}$, $\mathbf{X}$ and $\mathbf{X}_N$ be the stacked vectors of $\mathbf{h}_v$, $\mathbf{o}_v$, all features (node features, edge features, neighborhood features, etc.) and all the node features. Then the state embedding and output can be defined as:
\begin{align}
    \mathbf{H} &= F(\mathbf{H},\mathbf{X}), \\
    \mathbf{O} &= G(\mathbf{H},\mathbf{X}_N).
\end{align}

Due to the shallow learning mechanisms of most GNNs, one major problem of GNNs is the low expressive power limit. The main challenge of this problem is that most of the deep GNNs would suffer from the over-smoothing issue, i.e., the deep model would aggregate more and more node and edge information from neighbors which would lead to the representation of node and edge indistinguishable. Some works have been proposed to solve this problem recently. For example, in the work of \cite{DBLP:conf/nips/LuanZCP19}, the authors show that the Tanh activation function may be more suitable for deep GNNs and they also propose a DenseNet like architecture to alleviate the vanish-gradient problem.

To automate neural network exploration, some NAS methods have been proposed. Due to the substantial effort of human experts for discovering the state-of-the-art neural network architectures, there has been a growing interest in developing an automatic algorithm to design the neural network architecture automatically. Recently, the architectures generated by NAS have achieved state-of-the-art results in tasks like image classification, object detection or semantic segmentation. Most of the NAS methods are based on Reinforcement Learning (RL) \cite{DBLP:conf/iclr/ZophL17,DBLP:conf/cvpr/ZophVSL18,DBLP:conf/icml/PhamGZLD18} and Evolutionary Algorithm (EA) \cite{DBLP:conf/icml/RealMSSSTLK17,DBLP:conf/aaai/RealAHL19}. 

Although the aforementioned NAS methods have successfully designed CNN or RNN architectures for image and language modeling tasks, the GNN is very different from CNN or RNN. Thus they cannot be directly applied to the GNN architecture search. Gao et al. \cite{DBLP:journals/corr/abs-1904-09981} and Zhou et al. \cite{DBLP:journals/corr/abs-1909-03184} propose a new schema to encode the GNN architecture and apply reinforcement learning to search for GNN models, but their methods cannot generate deep GNNs and their methods are not efficient and robust enough.

\section{Method}

In this section, we first define the AutoGraph problem. Then we describe our search space and schema to represent GNN architectures. Next, we show our evolutionary algorithm for the AutoGraph. Finally, we show a method to improve the robustness of the search process.

\subsection{Problem Statement}
The AutoGraph problem can be formally defined as follows. Given search space $\mathcal{A}$, the target of our algorithm is to search the optimal GNN architecture $\alpha \in \mathcal{A}$ which minimizes the validation loss $\mathcal{L}_{val}$. It can be written as follows:
\begin{align}
    \text{min}_{\alpha}\quad &\mathcal{L}_{val}(w^*(\alpha),\alpha), \\
    \text{s.t. }\quad &w^{*}(\alpha) = \text{argmin}_w\ \mathcal{L}_{train}(w,\alpha),
\end{align}
where $w^{*}$ denotes the optimal parameters learned for the architecture in the training set. This is a bilevel optimization problem \cite{DBLP:journals/corr/abs-1904-09981}.

We propose an efficient method to solve this problem based on the evolutionary algorithm. Each generated architecture is trained and obtains the optimal weight of $w^*$ in the training set, then it is evaluated in the validation set. At last, the best architecture in the validation set is reported. The following sections explain the process in more detail.

\begin{figure}[tb]
    \centering
    \includegraphics[width=0.9\textwidth]{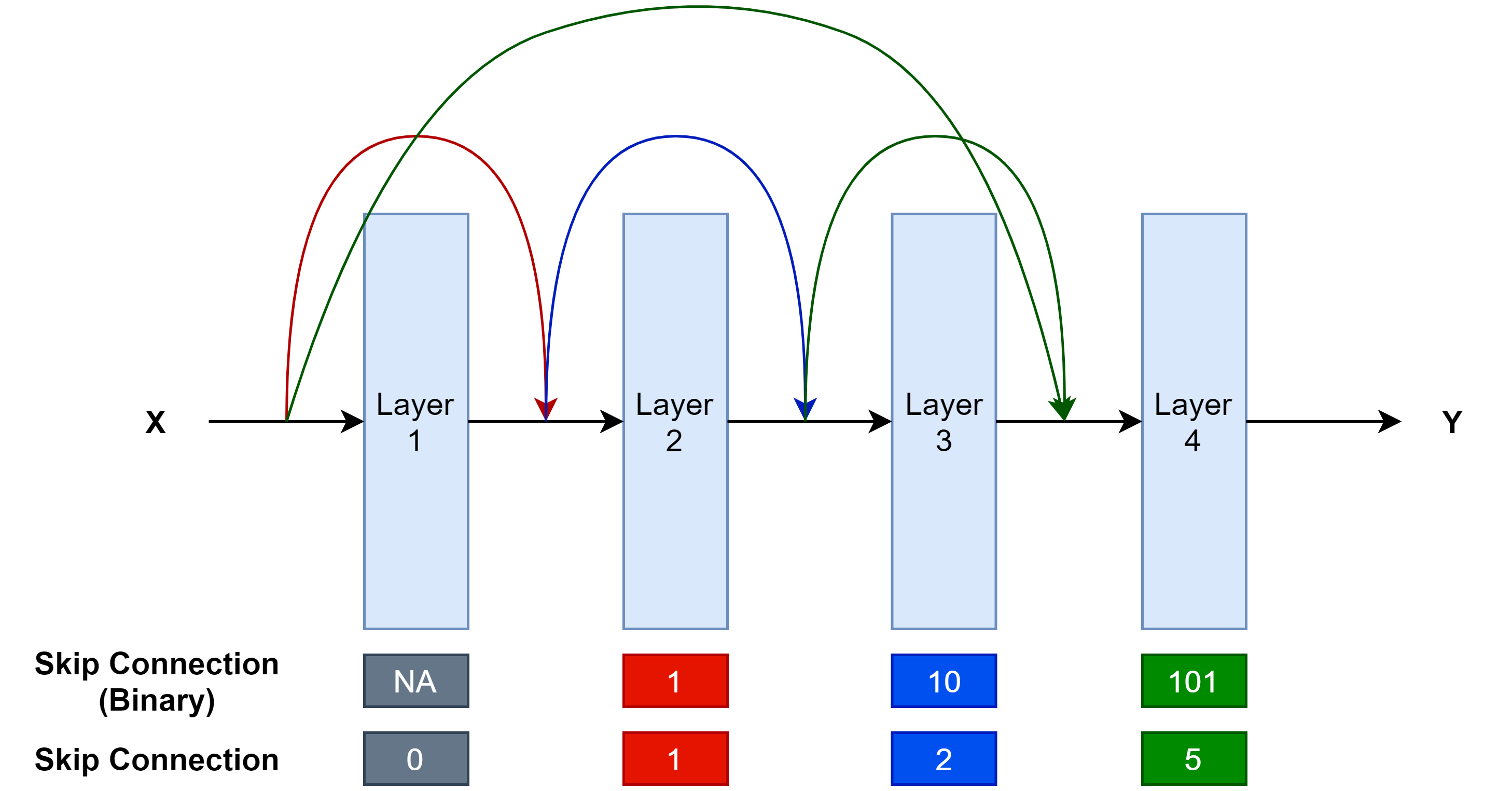}
    \caption{Graph skip connection example. Binary connection status can be encoded to [$0$,$2^{k-1}$), $k$ is the current layer number.}
    \label{fig:skipconnection}
\end{figure}

\subsection{Search Space}

Many state-of-the-art GNNs would suffer from the over-smoothing problem which makes the representation of even distant nodes indistinguishable \cite{DBLP:journals/corr/abs-1812-08434}. The recent work \cite{DBLP:conf/nips/LuanZCP19} shows that Tanh is better than ReLU for keeping linear independence among column features for GNNs. They propose a densely-connected graph network which is similar to DenseNet as follows:

\begin{align}
&\mathbf{H_0} = \mathbf{X}, \quad \mathbf{H_{l+1}} = f(L[\mathbf{H_0},\mathbf{H_1},...,\mathbf{H_l}]W_l), \quad l = 0,1,..,n-1, \\
&\mathbf{C} = g([\mathbf{H_0},\mathbf{H_1},..,\mathbf{H_n}]W_n), \\
&\mathrm{output} = \mathrm{softmax}(L^p\mathbf{C}W_C),
\end{align}
where $f$ and $g$ are activation functions; $W_l \in \mathbb{R}^{(\sum_{i=0}^{l}F_i)*F_{l+1}}$, $W_n \in \mathbb{R}^{(\sum_{i=0}^nF_i)*F_C}$ and $W_C \in \mathbb{R}^{F_C*F_O}$ are learnable parameters, $F_i$ is the number of input channels in layer $i$. This architecture stacks all the outputs of previous layers as the input of current layers. It can increase the variety of features for each layer, encourage the feature reuse, alleviate the vanishing gradient problem. However, concatenating all the outputs of previous layers will cause the parameters of the GNNs to increase exponentially.

Inspired by this, we allow each layer of our generated GNN models to connect to a various number of previous layers. To generate deep GNNs, we also allow our method to add a new layer to the GNN model during the searching process. So we define the search space and schema of our method as follows. We first apply the same setting of \textbf{Attention Function}, \textbf{Attention Head}, \textbf{Hidden Dimension}, \textbf{Aggregation Function} and \textbf{Activation Function} in \cite{DBLP:journals/corr/abs-1904-09981}. Then we introduce two new states:

\begin{itemize}
    \item \textbf{Skip Connection.} It has been observed that most GNN models deeper than two layers could not perform well because of the noisy information from expanding neighbors. This problem usually can be addressed by skip connection. Inspired by Luan et al. \cite{DBLP:conf/nips/LuanZCP19}, we allow skip connections between any previous layers to the current layer. For each previous layer, $0$ represents no skip connection, $1$ represents there is a skip connection between that layer to the current layer, e.g., Fig. \ref{fig:skipconnection}.
    \item \textbf{Layer Add\footnote{``Layer Add" state is only used in the evolutionary process}.} This state is only used during the mutation process. When this state is selected, we duplicate the current layer and add the new layer after the current layer. This state allows our method to extend the depth of GNNs automatically.
\end{itemize}

Noted that most of the GNN layers can be represented by the above first six states, as shown in Fig. \ref{fig:mutation}. The above search space can cover a wide variety of state-of-the-art GNN models. If the skip connections are applied then the input dimension of the current layer would be the sum of all the output dimensions of the connected layers.

\subsection{Evolutionary Algorithm}

Inspired by Real et al. \cite{DBLP:conf/aaai/RealAHL19}, we apply the Aging Evolution Algorithm to search for the deep GNNs. Similar to most of the evolutionary algorithms, our algorithm can be divided into three stages, i.e., initialization, mutation and updating. In the initialization stage, we randomly generate $P$ GNN models with two layers. $P$ is the size of the population. The initial $P$ models are trained and evaluated. Then they are added to the population.

In the mutation stage, we sample $S$ candidates from the population. The candidate with the highest score in the sample set is selected to apply mutation. We randomly select one state in the search space and change it to a new value in the state set. Then the newly generated candidate is trained and evaluated. Next, the new candidate needs to be added to the population. Since we need to keep the population size unchanged, we would select the oldest candidate in the population and remove it before we add the new candidate to the population. This is the main difference between the Aging Evolution Algorithm and other evolutionary algorithms.

We allow multiple skip connections for each layer. The skip connection between the previous layer $i$ to the current layer $k$ can be represented by binary $c_{i,k}$. Since there is always a connection between layer $k-1$ to layer $k$, we only need to consider $i \in 0,1,..,k-2$ ($0$ represents the input of the network). Thus, the skip connections state of layer $k$ can be represented as
\begin{equation}
    S_k = \sum_{i=0}^{k-2} c_{i,k} \cdot 2^i ,\quad c_{i,k} \in {0, 1}, \quad k >= 2.
\end{equation}
Then the possible state of $S_k$ is [$0$,$2^{k-1}$). When $k=1$, i.e., the current layer is the first layer, the skip connection state would be always $0$. Figure \ref{fig:skipconnection} shows an example of skip connection representation. To avoid a significant change of the GNN model, each mutation operation will only change one state of the model. During the search process, every evaluated GNN model is added to the history list. After the whole search process is finished, the model with the highest score in the history list will be reported.

\subsection{GNNs Evaluation}

We notice that the GNN model is sensitive to change in hyperparameters, such as the learning rate and weight decay. The best performance of a GNN architecture can be achieved at different learning rates, weight decay and iteration number. If we use the same hyperparameters to train and evaluate different GNN architectures, we may miss the best GNN model because the hyperparameters are not set properly. To fairly compare the architecture, we apply the hyperparameters tuning for each generated GNN model.

The work of Bergstra et al. \cite{DBLP:conf/nips/BergstraBBK11} shows that the  Tree-structured Parzen Estimator Approach (TPE) performs well on the hyperparameter search. We use the TPE algorithm to search the hyperparameters for each GNN model. To avoid overfitting and speed up the search process. We allow early stops during the training process. For each GNN architecture, we will use the best performance reported by the TPE algorithm as the performance of the architecture. The comparison between different GNN models is based on the performance of their best hyperparameter settings.

\section{Experiments}
We conduct experiments in both transductive and inductive learning tasks. For the transductive learning task, we test our method on the Cora, Citeseer and Pubmed datasets. For the inductive learning task, we test on the protein-protein interaction (PPI) dataset. Our method is evaluated in the following aspects:
\begin{itemize}
    \item \textbf{Performance.} We evaluate the performance of our AutoGraph method by comparing the generated GNN model with the handcrafted state-of-the-art GNN models.
    \item \textbf{Efficiency.} We analyze the efficiency of our method by comparing it with other search strategies, i.e., GraphNAS (a reinforcement learning-based method) and random search.
    \item \textbf{Scalability.} We analyze the scalability of our method by comparing the performance of GNN models with different layers.
\end{itemize}

\subsection{Experimental Setup}
 The configuration of our method in the experiments is set as follows. The population size is 100. The max evaluation architecture is 2,000. The maximum training iterations is 1,000. As described in the Methods, the mutation probabilities are uniform. The generated GNN architecture is trained with the ADAM optimizer. The maximum hyperparameters search number for the TPE algorithm is 50. We run the search algorithm in four RTX 2080 Ti GPU cards. For each task, the best model which has the lowest validation loss is selected as our GNN model to compare with other baseline models. 

\subsection{Datasets}

\textbf{Transductive Learning.} In transductive learning tasks, the same graphs are observed during training and testing. The experiment datasets for the transductive learning are Cora, Citeseer and Pubmed. In these datasets, the nodes represent the documents and the edges (undirected) represent citations. The features of the nodes are got by the bag-of-words representation of the documents. The Cora dataset contains 2,708 nodes and 5,429 edges. We will use 140 nodes for training, 500 nodes for validation and 1,000 nodes for testing. The Citeseer dataset contains 3,327 nodes and 4,732 edges. The training, validation and test set separations are the same as the setup of \cite{DBLP:conf/iclr/VelickovicCCRLB18}. 

\textbf{Inductive Learning.} In inductive learning tasks, the graphs in training and testing are different. The experiment dataset for inductive learning is the protein-protein interaction (PPI). The graphs in this dataset represent different human tissues. There are 20 graphs in the training set, two in the validation set and two in the test set. The data in the test set is completely unobserved during training. 

The statistical detail of transductive learning and inductive learning datasets is shown in Table \ref{tab:datasets}. The Cora, Citeseer and Pubmed datasets are classification problems. The PPI dataset is a multi-label problem.

\begin{table*}[tb]
\centering
\caption{Dataset Statistic}
\label{tab:datasets}
\begin{tabular}{l|c|c|c|c}
    \toprule
     &  \textbf{Cora} & \textbf{Citeseer} & \textbf{Pubmed} & \textbf{PPI}\\
     \midrule
    \textbf{Task} & \textit{Transductive} & \textit{Transductive} & \textit{Transductive} & \textit{Inductive} \\
    \midrule
    \textbf{\# Nodes} & 2,708 (1 graph) & 3,327 (1 graph) & 19,717 (1 graph) & 56,944 (24 graphs)\\
    \hline
    \textbf{\# Edges} & 5,429 & 4,732 & 44,338 & 818,716 \\
    \hline
    \textbf{\# Features/Node} & 1,433 & 3,703 & 500 & 50 \\
    \hline
    \textbf{\# Classes} & 7 & 6 & 3 & 121 (multi-label) \\
    \hline
    \textbf{\# Training Nodes} & 140 & 120 & 60 & 44,906 (20 graphs) \\
    \hline
    \textbf{\# Validation Nodes} & 500 & 500 & 500 & 6,514 (2 graphs) \\
    \hline
    \textbf{\# Test Nodes} & 1,000 & 1,000 & 1,000 & 5,524 (2 graphs) \\
    \bottomrule
\end{tabular}
\end{table*}

\subsection{Baseline Methods}

We compare the GNN model generated by our approach with the following state-of-the-arts methods: 
\begin{itemize}
    \item Chebyshev \cite{DBLP:conf/nips/DefferrardBV16}. This method removes the need to compute the eigenvectors of the Laplacian by using $K$-localized convolution to define a graph convolutional neural network.
    \item GCN \cite{DBLP:conf/iclr/KipfW17}. This method alleviates the problem of overfitting by limiting the layer-wise convolution operation to $K = 1$.
    \item GAT \cite{DBLP:conf/iclr/VelickovicCCRLB18}. This method introduces the attention mechanism to GNN. It obtains good results in many graph tasks.
    \item LGCN \cite{DBLP:conf/kdd/GaoWJ18}. It introduces regular convolutional operations to GNN.
    \item GraphSAGE \cite{DBLP:conf/nips/HamiltonYL17}. This method can be applied to inductive tasks. It samples and aggregates features from a node's neighborhood.
    \item GeniePath \cite{DBLP:conf/aaai/LiuCLZLSQ19}. It uses an adaptive path layer which consists of two complementary functions.
\end{itemize}

We use the public released implementations of these methods to do the comparisons. The evaluation metric for transductive learning tasks is accuracy. For the inductive learning task, we use the micro-F1 score.

To evaluate the efficiency of our method, we also compare our method with GraphNAS and random search. GraphNAS applies a reinforcement learning controller to generate GNN models. For the random search baseline, we randomly sample GNN models from the same search space in our approach.

\begin{table}[t]
    \centering
    \caption{Experiment results on Cora, Citeseer and Pubmed}
    \label{tab:tranductive_result}
    \begin{tabular}{>{\centering\arraybackslash}m{3cm}| >{\centering\arraybackslash}m{3cm}| >{\centering\arraybackslash}m{3cm}| >{\centering\arraybackslash}m{3cm}}
    \toprule
    \textbf{Models} & \textbf{Cora} & \textbf{Citeseer} & \textbf{Pubmed} \\
    \midrule
    Chebyshev & $81.2\%$ & $69.8\%$ & $74.4\%$ \\
    GCN & $81.5\%$ & $70.3\%$ & $79.0\%$ \\
    GAT & $83.0\pm 0.7\%$ & $72.5\pm 0.7\%$ & $79.0\pm 0.3\%$ \\
    LGCN & $83.3\pm 0.5\%$ & $73.0\pm 0.6\%$ & $79.5\pm 0.2\%$ \\
    GraphNAS & $83.3\pm 0.6\%$ & $73.5\pm 1.0\%$ & $78.8\pm 0.5\%$ \\
    \hline
    AutoGraph & $\bm{83.5\pm 0.4\%}$ & \bm{$74.4\pm 0.4\%$} & $\bm{80.3\pm 0.3\%}$ \\
    \bottomrule
    \end{tabular}
\end{table}

\begin{table}[tbp]
    \centering
    \caption{Experiment results on PPI}
    \label{tab:inductive_result}
    \begin{tabular}{>{\centering\arraybackslash}m{3cm}| >{\centering\arraybackslash}m{3cm}}
    \toprule
    \textbf{Models} & \textbf{micro-F1} \\
    \midrule
    GraphSAGE (lstm) & $0.612$ \\
    GeniePath & $0.979$ \\
    GAT & $0.973\pm 0.002$ \\
    LGCN & $0.772\pm 0.002$ \\
    GraphNAS & $0.985\pm 0.004$ \\
    \hline
    AutoGraph & \bm{$0.987 \pm 0.003$}  \\
    \bottomrule
    \end{tabular}
\end{table}



\subsection{Results}

After our algorithm generates 2,000 GNN models, the model which has the lowest loss in the validation set is selected and tested on the test set. The experiment results of transductive learning datasets are summarized in Table \ref{tab:tranductive_result}. The results of the inductive learning dataset are summarized in Table \ref{tab:inductive_result}. 


\textbf{Performance.}
For the transductive learning tasks, we compare the classification accuracy with the above-mentioned GNN model and GraphNAS. From Table \ref{tab:tranductive_result} we can see that our generated model can get the state-of-the-art result in all transductive datasets.

For the inductive task, we compare the micro-F1 score with the popular GNN models and GraphNAS. The result shows that our method also performs well in the inductive dataset.

\begin{table}[bt]
    \centering
    \caption{Search Strategies Comparison}
    \label{tab:search_strategies}
    \begin{tabular}{>{\centering\arraybackslash}m{3cm}| >{\centering\arraybackslash}m{3cm}| >{\centering\arraybackslash}m{3cm}| >{\centering\arraybackslash}m{3cm}}
    \toprule
    \textbf{Method} & \textbf{Accuracy} & \textbf{Time (GPU hours)} & \textbf{Best GNN Layers} \\
    \midrule
    Random Search & $81.8 \pm 0.5\%$ & 10 & 2\\
    GraphNAS & $83.3 \pm 0.6\%$ & 10 & 2 \\
    \hline
    AutoGraph & $\bm{83.5 \pm 0.4\%}$ & \bm{$3$} & \bm{$4$}  \\
    \bottomrule
    \end{tabular}
\end{table}

\textbf{Efficiency.}
To evaluate the effectiveness of our search method, we compare our method with different search strategies, i.e., random search and reinforcement learning-based search method\textemdash GraphNAS \cite{DBLP:journals/corr/abs-1904-09981}. Since GraphNAS does not do the hyperparameters tuning when evaluating the GNNs, we also disable our hyperparameters tuning during the search process. During the training process, we record the generated architectures and their performance. From the Table \ref{tab:search_strategies}, we can see that our method can search for a better GNN model with less time and our method can generate deeper GNNs.

\begin{figure}[tb]
    \centering
    \includegraphics[width=0.5\textwidth]{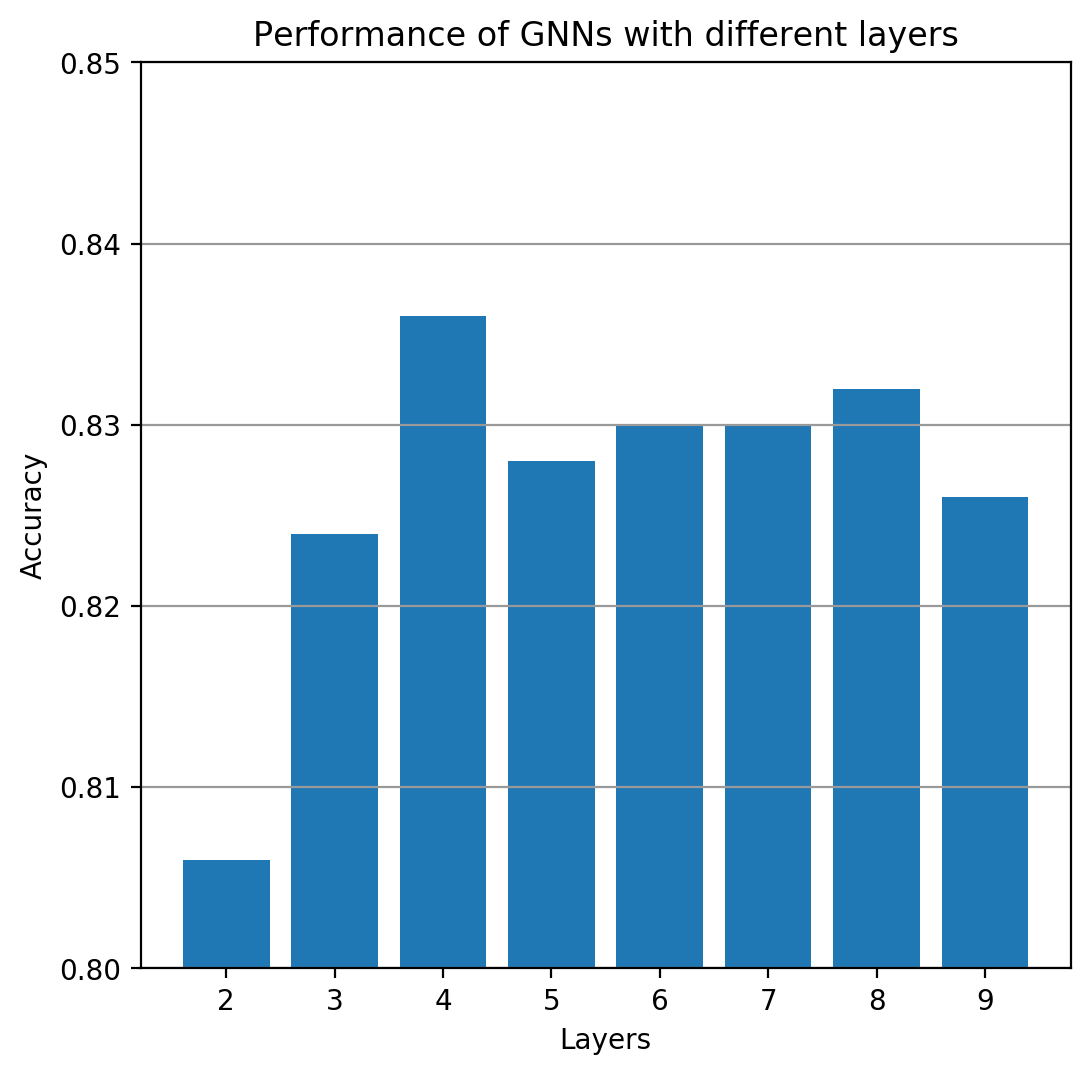}
    \caption{Comparison of the GNN models with different layers on Cora}
    \label{fig:layers_comparison}
\end{figure}

\textbf{Scalability.}
We know that most of the handcraft GNNs would suffer from the over-smoothing problem. We compare the performance of the GNNs generated by our method with different layers. Figure \ref{fig:layers_comparison} shows the best performance of the GNNs generated by our method from two layers to nine layers. We can see that our generated GNN models have good performance in deep architectures.

\section{Discussion \& Conclusion}

In this work, we study the problem of AutoGraph. We present an efficient evolutionary algorithm to search for GNN models. We can see that our method can generate deep GNNs which alleviate the over-smoothing problem. The experiments show that the generated models can outperform current handcraft state-of-the-art models. In summary, we can see our proposed method has the following advantages:
\begin{itemize}
    \item It can save substantial efforts to explore good GNN models for different graph tasks.
    \item Our generated GNN models can get state-of-the-art results.
    \item Our approach can generate deep GNN models which can alleviate the over-smoothing problem.
\end{itemize}

Although our proposed method can design state-of-the-art GNNs for graph tasks, it is remarkable that there are still many improvements that can be made. The first problem is that the search process is time-consuming. We notice that some approaches to reduce the search time have been proposed in NAS for CNNs. However, most of them cannot be directly applied to GNNs, we need to design a proper improvement method for GNNs. The second problem is that the search space in our method is still limited, we can try to design a better search space to explore more novel GNNs. We will focus on these two problems in our future works.


%
%

\bibliographystyle{splncs04}

\end{document}